\documentclass[10pt,twocolumn,letterpaper,dvipsnames,table]{article}

\usepackage{cvpr}              

\definecolor{cvprblue}{rgb}{0.21,0.49,0.74}
\usepackage[pagebackref,breaklinks,colorlinks,allcolors=cvprblue]{hyperref}
\usepackage{graphicx}
\usepackage{adjustbox}
\def\paperID{50} 
\def\confName{CVPR}
\def\confYear{2025}

\title{RareSpot: Spotting Small and Rare Wildlife in Aerial Imagery with Multi-Scale Consistency and Context-Aware Augmentation}

\author{
Bowen Zhang$^{1}$ \quad
Jesse T. Boulerice$^{2}$ \quad
Nikhil Kuniyil$^{1}$ \quad
Charvi Mendiratta$^{1}$ \quad \\
Satish Kumar$^{3}$ \quad
Hila Shamon$^{2}$ \quad
B.S. Manjunath$^{1}$ \\
$^{1}$University of California, Santa Barbara \\ 
$^{2}$Smithsonian National Zoo and Conservation Biology Institute; 
$^{3}$Stanford University 
}

\newcommand{\ignore}[1]{}

\begin{document}
\maketitle

\begin{abstract}
\vspace{-5pt}

Automated detection of small and rare wildlife in aerial imagery is crucial for effective conservation, yet remains a significant technical challenge. Prairie dogs exemplify this issue: their ecological importance as keystone species contrasts sharply with their elusive presence—marked by small size, sparse distribution, and subtle visual features—which undermines existing detection approaches. To address these challenges, we propose \textbf{RareSpot}, a robust detection framework integrating multi-scale consistency learning and context-aware augmentation. Our multi-scale consistency approach leverages structured alignment across feature pyramids, enhancing fine-grained object representation and mitigating scale-related feature loss. Complementarily, context-aware augmentation strategically synthesizes challenging training instances by embedding difficult-to-detect samples into realistic environmental contexts, significantly boosting model precision and recall. Evaluated on an expert-annotated prairie dog drone imagery benchmark, our method achieves state-of-the-art performance, improving detection accuracy by over 35\% compared to baseline methods. Importantly, it generalizes effectively across additional wildlife datasets, demonstrating broad applicability. The \textit{RareSpot} benchmark and approach not only support critical ecological monitoring but also establish a new foundation for detecting small, rare species in complex aerial scenes.



\end{abstract}    
\section{Introduction}
\begin{figure}[t]
    \centering
    \includegraphics[width=\linewidth]{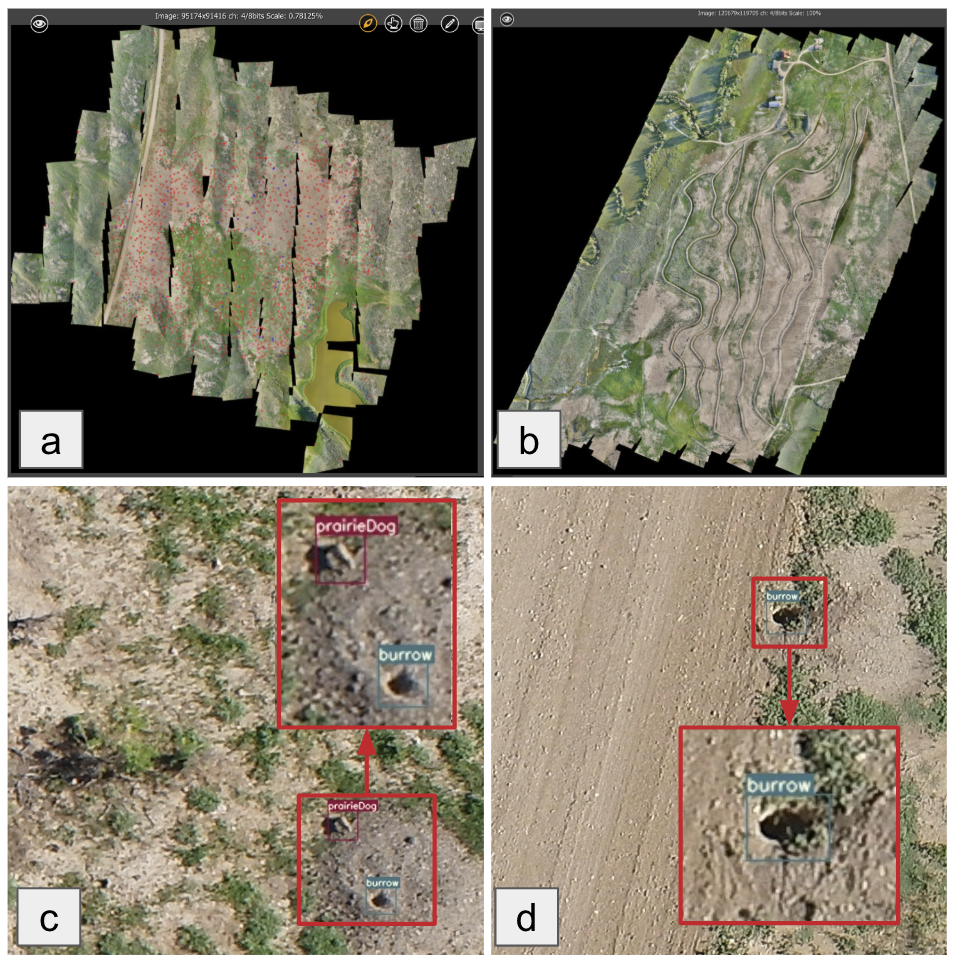}
    \caption{ \it Overview of our Prairie Dog Drone dataset. Drone orthomosaics in (a) and (b) illustrate large-scale drone deployment (covering approximately 2 km\textsuperscript{2}). Examples in (c) and (d) show detected prairie dogs and burrows. Insets provide zoomed-in views to emphasize the difficulty of finding objects from natural background textures such as soil, vegetation, or shadows.}
    \label{fig:data}
\end{figure}
\hspace{1em} 
Many studies have explored automated wildlife monitoring using computer vision and deep learning ~\cite{tuia2022perspectives, mcever2023context, kellenberger2018detecting, delplanque2023crowd}. However, current detectors still face poor performance on small (less than 30 pixel in length) and rare (occupying less than 0.1\% of the whole image) animal species, especially under real-world conditions. Prairie dogs exemplify this challenge. Although ecologically vital as a keystone species~\cite{Koltiar,boulerice2019use}, they appear in drone images as extremely small, low-contrast objects often camouflaged by natural backgrounds~\cite{axford2024collectively}. Burrow entrances are similarly subtle and often confused with shadows or rocks. These combined challenges make it difficult for state-of-the-art detectors or naively scaled-up models to learn the necessary fine-grained features. Notably, simply applying larger, general-purpose vision models does not effectively solve these issues~\cite{rekavandi2023transformers}.

\begin{figure*}[t]
    \centering
    \includegraphics[width=0.85\linewidth]{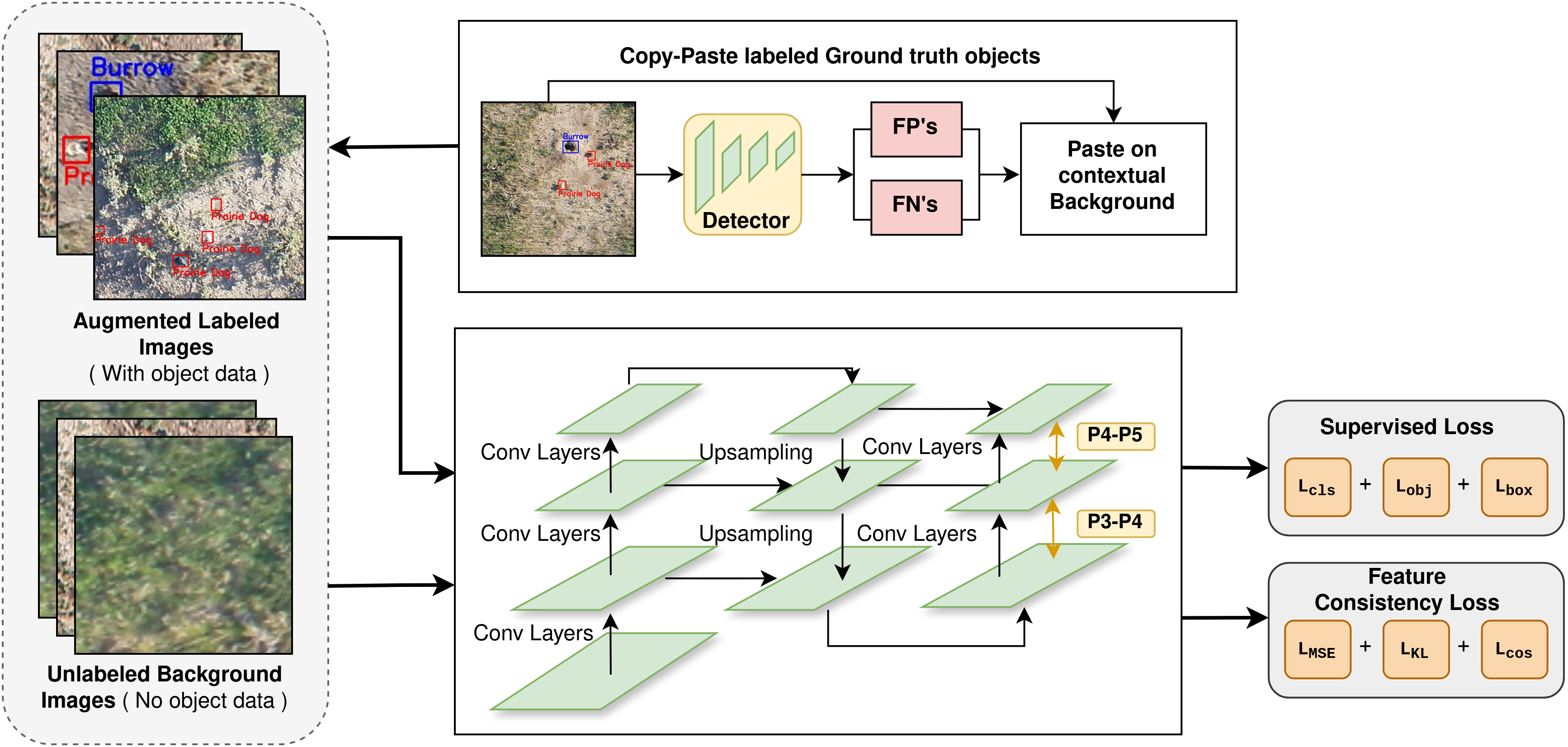} 
    \caption{\it Overview of all feature layers with the proposed Feature Consistency Losses and Context-Aware Hard Sample Augmentation.}
    \label{fig:pipeline}
\end{figure*}

In this work, we introduce a targeted framework designed to address the unique challenges of detecting small and rare wildlife in aerial imagery. Our key contributions include:


\begin{itemize}
    \item \textbf{Multi-Scale Consistency Learning}: We propose three complementary consistency losses applied across feature pyramid scales, preserving fine-grained details in higher-resolution features while guiding lower-resolution representations. This significantly enhances the detection of small objects, improving recall and reducing false positives.


     \item \textbf{Context-Aware Hard Sample Augmentation}: We develop a targeted augmentation strategy that identifies model failure cases, including false positives and false negatives, and embeds them into contextually appropriate backgrounds. This process addresses data scarcity and improves the model’s ability to generalize to cluttered and visually ambiguous natural scenes.

    \item \textbf{Expert-Annotated Prairie Dog Dataset}: We introduce the first expert-validated drone imagery dataset of prairie dogs and burrows, providing a robust benchmark for advancing small-object wildlife detection research.

\end{itemize}

\begin{table*}[t]
\centering
{\footnotesize

    \caption{\it Statistical summary of the prairie dog habitat datasets used in this study. The table presents the distribution of annotated images, tiles, prairie dog (PD) burrows, and orthomosaic dimensions across training, validation, and test splits.}
    \label{tab:data_split}
    \renewcommand{\arraystretch}{1} 
    \setlength{\tabcolsep}{8pt} 
    \rowcolors{2}{gray!15}{white}
    \begin{tabular}{lcccccc} 
        \toprule
        \textit{Dataset} & \textit{Split} & \textit{\# of annot. images} & \textit{\# of annot. tiles} & \textit{\# of PDs} & \textit{\# of Burrows} & \textit{Orthomosaic Size} \\ 
        \midrule
        EnrNE\_d2\_r1  & Train & 328 & 7,858 & 807 & 3913 & 95K$\times$91K \\
        EnrNE\_d2\_r2 & Validation & 329 & 9,293 & 569 & 4435 & 85K$\times$84K \\
        Enr\_d1\_r1 & Test & 739 & 7,616 & 474 & 3642 & 57K$\times$73K \\
        \bottomrule
    \end{tabular}	
}
\end{table*}


\section{Datasets and Methods}
\label{sec:methods}


\hspace{1em} Multiscale feature processing is well-studied \cite{huang2022small,huynh2024one}. {\em However, our novel approach uniquely enforces fine-grained consistency across scales via multiple complementary feature-level losses}. Figure~\ref{fig:pipeline} illustrates our pipeline, comprising a backbone, multi-scale detection head, and context-aware augmentation.

\subsection{Datasets}
Our Prairie Dog datasets comprise aerial imagery from two distinct prairie dog colonies covering approximately 2 km\textsuperscript{2}, captured by a fixed-wing drone at 100m altitude with 2cm/pixel resolution. Images were mosaicked from flights using 70\% overlap and annotated for two classes (``prairie dog" and ``burrow") by trained ecology volunteers, with expert validation. Dataset statistics are detailed in Table~\ref{tab:data_split}.
{To prevent spatial leakage, we split the dataset by drone flight. Training and validation sets are from Colony A but captured on different dates (328 images/15,893 tiles for training; 329 images/4,788 tiles for validation, shown in Figure \ref{fig:data}a), while the test set is from a separate survey over Colony B survey (739 images/7,616 tiles, shown in Figure \ref{fig:data}b). Prairie dog bounding boxes range from 11–98 pixels in width (mean 33.5) and  10–129 pixels in height (mean 33.4), reflecting their small size. Burrow boxes are slightly larger and more variable, with widths of  19–210 pixels in width (mean 52.4) and heights of 16–200 pixels (mean 50.7). Each 512×512 tile contains on average 0.07 prairie dogs and 0.48 burrows. In total, the dataset includes  consists of 24,767 image tiles from 1,396 drone images, with 1,850 prairie dogs and 11,900 burrows annotated.}


\subsection{Multi-Scale Consistency Learning}

Effective small-object detection demands feature consistency across scales. Our YOLOv5 backbone employs an FPN+PAN architecture producing feature maps at three scales ($P_3, P_4, P_5$), balancing fine-grained details ($P_3$) with broader context ($P_4, P_5$). Misalignments between scales degrade detection accuracy. We propose \textit{multi-scale consistency learning} to enforce structured alignment across these scales.


\subsection{Feature Alignment Across Scales}

To ensure consistency in multi-scale detection, we enforce structured alignment between feature maps at different resolutions. Given three feature maps extracted by YOLOv5’s FPN + PAN architecture,

{\footnotesize
\begin{equation}
P_3 \in \mathbb{R}^{C \times H \times W}, 
P_4 \in \mathbb{R}^{C \times \frac{H}{2} \times \frac{W}{2}}, 
P_5 \in \mathbb{R}^{C \times \frac{H}{4} \times \frac{W}{4}}.
\end{equation}
}

\noindent where \(C\) is the number of channels and \((H, W)\) is the spatial dimension of \(P_3\), we upsample \(P_4\) and \(P_5\) to match the highest resolution, so that
\begin{align}
    \tilde{P}_4  \in \mathbb{R}^{C \times H \times W}, 
    \tilde{P}_5  \in \mathbb{R}^{C \times H \times W}.
\end{align}


\noindent
\textbf{Mean Squared Error (MSE):}
MSE enforces similarity in raw activation magnitudes. For each pixel \((i, j)\), we compute the squared difference between feature vectors:

{\footnotesize
\begin{align}
\mathcal{L}_{\mathrm{MSE}} = 
\frac{1}{HW} \sum_{i=1}^H \sum_{j=1}^W 
\Bigl\| P_3^{(i,j)} - \tilde{P}_4^{(i,j)} \Bigr\|^2 
+ \Bigl\| \tilde{P}_4^{(i,j)} - \tilde{P}_5 ^{(i,j)}\Bigr\|^2
\end{align}
}

This constraint ensures that coarse-scale feature vectors retain numerical similarity to their high-resolution counterparts, preserving \textit{small-object details} across the feature hierarchy.

\noindent
\textbf{Kullback--Leibler (KL) Divergence:}
To complement MSE, which maintains numerical consistency, we introduce a KL divergence-based alignment loss that ensures semantic coherence across scales. Treating feature vectors as probability distributions via softmax normalization, the KL divergence loss is defined as:

{\footnotesize
\begin{equation}
\begin{aligned}
L_{\mathrm{KL}} = 
\frac{1}{HW} \sum_{i=1}^H \sum_{j=1}^W 
\mathrm{KL}\bigl(S[P_3]^{(i,j)},\, S[\tilde{P}_4]^{(i,j)}\bigr) \\
+ \mathrm{KL}\bigl(S[\tilde{P}_4]^{(i,j)},\, S[\tilde{P}_5]^{(i,j)}\bigr)
\end{aligned}
\end{equation}
}

\noindent where $\mathrm{KL}(\cdot,\cdot)$ denotes KL divergence and $S[\cdot]$ is the softmax function.

\noindent
\textbf{Cosine Similarity:} Unlike MSE and KL divergence, cosine similarity loss maintains angular relationships between feature vectors, ensuring consistent orientation across scales even when small-object features diminish. The loss is defined as:

{\footnotesize
\begin{equation}
\begin{aligned}
\mathcal{L}_{\mathrm{cos}} = 
\frac{1}{HW} \sum_{i=1}^H \sum_{j=1}^W 
\Bigl[1 - \cos\bigl(P_3^{(i,j)},\, \tilde{P}_4^{(i,j)}\bigr)\Bigr] \\
+ \Bigl[1 - \cos\bigl(P_3^{(i,j)},\, \tilde{P}_5^{(i,j)}\bigr)\Bigr].
\end{aligned}
\end{equation}
}

\subsection{Final Multi-Scale Consistency Loss}
To enforce coherent feature representation across scales, we define our multi-scale consistency loss as:

{\footnotesize
\begin{equation}
\mathcal{L}_{\mathrm{consistency}} 
\;=\;
\alpha \,\mathcal{L}_{\mathrm{MSE}}
\;+\;
\beta \,\mathcal{L}_{\mathrm{KL}}
\;+\;
\gamma \,\mathcal{L}_{\mathrm{cos}},
\end{equation}
}

\noindent where \(\mathcal{L}_{\mathrm{MSE}}\) reduces numerical divergence, \(\mathcal{L}_{\mathrm{KL}}\) aligns semantic distributions, and \(\mathcal{L}_{\mathrm{cos}}\) preserves angular consistency. This combined loss ensures small-object features remain robustly detectable at all scales, significantly improving performance in complex imagery.


In our fixed-altitude aerial survey setting, targets appear at a consistently small and uniform scale. Our method leverages this by enforcing cross-scale feature coherence tailored for small-object detection. Each of the three alignment losses is assigned an independent weight, allowing fine-grained control over their influence and selective application to specific pyramid levels. This preserves P3’s high-resolution detail and P5’s semantic context, maintaining each level’s scale-specific role while improving overall feature robustness. The consistency losses are added on top of the YOLO detection loss with tunable weights, preserving the model’s ability to detect objects across scales.

\subsection{Context-Aware Hard Sample Augmentation for Rare and Small Object Detection}
\begin{table*}[t]

\centering
\caption{\it Performance comparison of different model variations on the validation and test sets for prairie dog and burrow detection. The proposed \textit{RareSpot}, which integrates Multi-Scale Consistency Learning and Context-Aware Hard Sample Augmentation, outperforms the baseline across both datasets. These results demonstrate the effectiveness of incorporating cross-scale feature consistency and hard sample prioritization in improving small-object detection.}
\label{tab:results}
\resizebox{\textwidth}{!}{
{
\begin{tabular}{l|cc|cc|cc|c}
\hline
Model & \multicolumn{2}{c|}{Precision (P)} & \multicolumn{2}{c|}{Recall (R)} & \multicolumn{2}{c|}{mAP@50} & Overall mAP@50 \\
 & PD & Burrow & PD & Burrow & PD & Burrow & \\
\hline
\multicolumn{8}{c}{Validation Results} \\
\hline
Baseline Model (YOLOv5L) & 0.570 & 0.791 & 0.370 & 0.840 & 0.393 & 0.846 & 0.620 \\

Multi-Scale Consistency Learning & 0.578 & 0.824 & 0.434 & 0.850 & 0.432 & 0.843 & 0.637 \\
CA Hard Sample Augmentation & 0.557 & 0.823 & 0.407 & 0.851 & 0.415 & 0.852 & 0.633 \\
\textbf{RareSpot (Combined)} & \textbf{0.634} & \textbf{0.845} & \textbf{0.470} & \textbf{0.885} & \textbf{0.491} & \textbf{0.871} & \textbf{0.681} \\
\hline
Improvement over Baseline & \textbf{+11.23\%} & \textbf{+6.83\%} & \textbf{+27.03\%} & \textbf{+5.36\%} & \textbf{+24.94\%} & \textbf{+2.96\%} & \textbf{+9.90\%} \\
\hline
\multicolumn{8}{c}{Test Results} \\
\hline
Baseline Model (YOLOv5L) & 0.482 & 0.790 & 0.366 & 0.891 & 0.366 & 0.886 & 0.626 \\

Multi-Scale Consistency Learning & 0.536 & 0.842 & 0.488 & 0.910 & 0.454 & 0.914 & 0.684 \\
CA Hard Sample Augmentation & 0.464 & 0.871 & 0.471 & 0.891 & 0.383 & 0.912 & 0.647 \\
\textbf{RareSpot (Combined)} & \textbf{0.591} & \textbf{0.893} & \textbf{0.519} & \textbf{0.907} & \textbf{0.495} & \textbf{0.923} & \textbf{0.709} \\
\hline
Improvement over Baseline & \textbf{+22.61\%} & \textbf{+13.04\%} & \textbf{+41.80\%} & \textbf{+1.80\%} & \textbf{+35.25\%} & \textbf{+4.18\%} & \textbf{+13.26\%} \\
\hline
\end{tabular}
}
}
\end{table*}

Prairie dogs exemplify an extreme small-object detection challenge in drone imagery: they comprise only $\sim 0.8\%$ of our dataset and exhibit weaker visual signals compared to large-scale benchmarks (e.g., COCO). To address this, we introduce a \textit{context-aware hard sample augmentation} strategy that leverages detection failures and environmental priors to improve robustness and generalization.

Our approach begins by mining hard examples from previous model outputs, specifically false positives (FPs) and false negatives (FNs), which often correspond to ambiguous visual patterns such as shadows, rocks, or vegetation textures. These are supplemented by true labeled instances. We extract image patches centered on these objects and embed them into \textit{semantically consistent} backgrounds to create challenging yet realistic training examples.

To construct background images, we select an equal number of empty (background-only) images from the training set and perform HSV color-based segmentation to identify habitat sub-regions, namely: dirt/mud and grass. This segmentation defines the \emph{context map} \( c \), where \( c(i, j) \in \{\text{dirt}, \text{grass}\} \) labels each pixel as belonging to a specific habitat type.

Each patch from \( P_{\text{Labeled}}, P_{\text{FP}}, P_{\text{FN}} \) is then assigned to a contextually appropriate region: 90\% of the patches are placed in dirt regions and 10\% in grass regions, reflecting their distribution in real habitats. Prior to placement, each patch undergoes a sequence of transformations defined by parameter set \( \theta = \{\theta_{\text{scale}}, \theta_{\text{rot}}, \theta_{\text{illum}}\} \), 
which include:

\begin{itemize}
    \item $\theta_{\text{scale}}$: random scaling in the range $[0.9, 1.1]$
    \item $\theta_{\text{rot}}$: random in-plane rotation in the range $[-90^\circ, +90^\circ]$
    \item $\theta_{\text{illum}}$: random brightness and contrast adjustments (HSV adjustment)
\end{itemize}

These augmented patches are blended into the background using  Poisson image blending, via SeamlessClone library from OpenCV\cite{opencv_library}. 

Formally, the augmentation process is defined as:
{\footnotesize
\begin{equation}
I' \;=\; \mathrm{Aug}\!\Bigl(
   I,\;
   \{P_{\text{FPs}},\,P_{\text{FNs}},\,P_{\text{Labeled}}\}\;   ;\;
   c,\; \theta
\Bigr),
\end{equation}
}

\noindent where $I$ is a background image, \( \{P_{\text{FP}}, P_{\text{FN}}, P_{\text{Labeled}}\} \) are the sets of extracted hard and ground-truth patches, \( c \) is the pixel-level context map identifying habitat type, and \( \theta \) is the set of augmentation parameters applied to each patch before placement. The output $I'$ represents the final augmented image, generated by embedding the transformed patches into semantically consistent regions of the background image I, as guided by the context map c.

This process strategically increases the occurrence of challenging instances while ensuring ecological realism. It improves the model’s ability to distinguish small prairie dogs and burrows from visually similar clutter, thereby enhancing both recall and precision in real-world deployment. In total, the augmentation pipeline generated 3,929 additional images, which were incorporated into the training set to strengthen representation of rare and ambiguous cases.

\section{Experiments}
\label{sec:experiments}

\hspace{1em} Table~\ref{tab:results} presents our detection performance for prairie dogs and burrows on both validation and test sets. The \textbf{Baseline Model} (YOLOv5-L with high augmentation settings) achieves strong burrow detection but struggles with the smaller, low-contrast prairie dogs.

\noindent \textbf{Multi-Scale Consistency Learning} enhances detection performance for prairie dogs by reinforcing their feature representation across multiple spatial resolutions. On the validation set, this approach improves prairie dog mAP by 17.3\%. The improvement is further validated on the test set, where prairie dog mAP@50 rises by 24\%, highlighting the method’s effectiveness in capturing fine-grained, discriminative features for small-object detection.
\noindent \textbf{Context-Aware Hard Sample Augmentation} adds more labels on different background, focusing on challenging instances. Our augmentation approach raises recall and precision for both objects in all datasets. 
\noindent \textbf{Our RareSpot Model} integrates above approaches and delivers the strongest overall performance. Specifically, it increases prairie dog mAP@50 by 35.25\% and recall by 41.80\% on the test set. These improvements clearly demonstrate the synergy of our integrated method, enabling robust detection of prairie dogs.


\noindent \textbf{Choice of Backbone and Comparison with State-of-the-Art Methods:}
We adopt YOLOv5 as our base detector due to its strong performance on small-object detection and its ability to preserve fine-grained features through static feature aggregation and fixed anchors. In our comparisons (Tables~\ref{tab:results} and~\ref{tab:othermodels}), other FPN-based backbones—including ResNet-50-FPN, YOLOv7, YOLOv10, and ViT-FPN—consistently underperformed relative to YOLOv5 on our target datasets. As such, we focused on enhancing the most competitive baseline.

Although YOLOv7 shares the CSPDarknet-FPN backbone, its architectural changes (RepConv, E-ELAN, dynamic label assignment) prioritize medium-to-large objects and tend to degrade high-resolution features crucial for detecting small targets~\cite{li2024YOLOUAV, li2023improved}. Our Multi-Scale Consistency Learning module integrates seamlessly with YOLOv5, introducing no additional parameters or dataset-specific tuning, yet yielding consistent mAP improvements—even on mixed-scale datasets like WAID (object sizes: 20–530 pixels). 

As shown in Table~\ref{tab:othermodels}, our combined approach outperforms state-of-the-art detectors, including transformer-based models like DETR~\cite{carion2020end}, Co-DETR, and WildlifeMapper, as well as specialized architectures like TPH-YOLO. Although these models excel on large-scale datasets, their reliance on abundant training data reduces their effectiveness in scenarios like ours, which involve limited annotations and rare, small-sized targets.



 \begin{figure}[t]
    \centering
    \includegraphics[width=\linewidth]{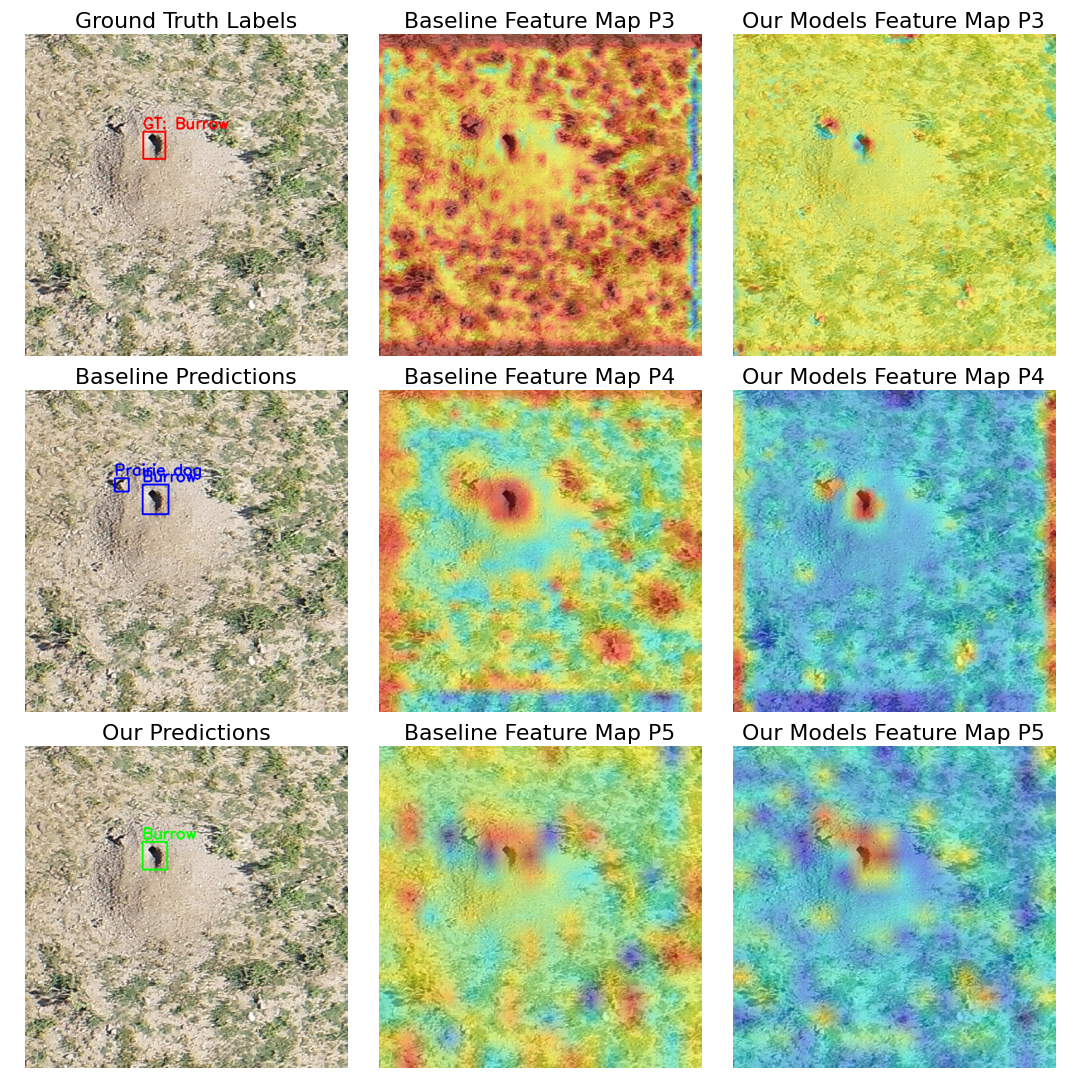}
    \caption{\it Feature map visualization for P3, P4, and P5. Images in the first column are ground truth, baseline predictions, and Ours predictions with the threshold of 0.2. The last 2 columns are features maps for the baseline and our model.}
    \label{fig:feature_map}
\end{figure}

\begin{table}[t]
\centering
{
\caption{\it Comparison with other detection models on the validation set.}
\label{tab:othermodels}

\begin{tabular}{l|c}
\hline
\textbf{Models} & \textbf{mAP@50} \\
\hline
Baseline Model (YOLOv5L) & 0.620 \\
YOLOv5L (P3 Only) & 0.623 \\
YOLOv7~\cite{wang2022yolov7} & 0.517 \\
YOLOv10~\cite{chen2024yolov10} & 0.471 \\
DETR~\cite{carion2020end} & 0.411 \\
Co-DETR~\cite{zong2023detrs} & 0.437 \\
TPH-YOLO~\cite{zhu2021tph} & 0.534 \\
WildlifeMapper~\cite{kumar2024wildlifemapper} & 0.313 \\
\textbf{RareSpot} & \textbf{0.681} \\
\hline
\end{tabular}
}
\end{table}

\noindent \textbf{Performance on Other Drone-Based Animal Datasets:} 
To assess the generalization of our Multi-Scale Consistency Learning, we applied the same pipeline—without dataset-specific tuning or additional annotations—to three diverse drone-based wildlife benchmarks: the Aerial Elephant Dataset~\cite{Naude_2019_CVPR_Workshops}, the Waterfowl Thermal Imagery Dataset~\cite{Huwaterfowl2024}, and WAID~\cite{mou2023waid}. Despite differences in modality (visible vs. thermal) and species, our method consistently outperformed the YOLOv5 baseline, achieving +4.5\% mAP on elephants and +3.9\% on waterfowl. On WAID, it reached 97.6\% mAP—just 0.7\% below SE-YOLO~\cite{mou2023waid}, the best-reported result, though that method lacks public code or pretrained weights, limiting reproducibility. The narrow margin further underscores the competitiveness and broad applicability of our approach.

\begin{table}[h]
{
    \centering
    \begin{tabular}{lccc}
        \hline
        Method & AED & Waterfowl Thermal & WAID\\
        \hline
        YOLOv5 & 0.891 & 0.957 & 0.956\\
        WM & 0.553 & 0.368 & 0.658\\
        \textbf{Ours} & \textbf{0.897} & \textbf{0.970} & \textbf{0.976} \\
        \hline
    \end{tabular}
    \caption{\it Performance comparison of different methods on other small scale datasets. WM stands for WildlifeMapper.}
    \label{tab:otherdatasets}
}
\end{table}

\noindent\textbf{Qualitative Analysis}
Qualitative results demonstrate that our method consistently localizes prairie dogs and burrows more accurately than the baseline model. The detections and feature maps are more precise and less susceptible to noise. These observations further validate the effectiveness of our combined approach and underscore its suitability for real-world deployment in complex environmental settings. We provide five detailed figures and analysis in Section~\ref{sec:sup_qual} of the Supplementary Materials.

\section{Discussion} Our detection pipeline excels at spotting small, rare wildlife in high-resolution aerial imagery by combining multi-scale feature consistency and context-aware augmentation, yielding substantial gains over both baselines and large transformer-based methods on our prairie dog drone dataset. Looking ahead, we plan to incorporate abundant unlabeled and background-only data via semi-supervised learning, aiming to enhance generalization in data-scarce settings, reduce annotation costs, and further improve ecological monitoring performance in real-world applications.

\section{Acknowledgement}
This research was supported in part by the the NSF CSSI Award \#2411453. We thank the UCSB BisQue team: Chandrakanth Gudavalli, Connor Leverson, Amil Khan, and Umang Garg, for their technical support, and Lacey Hughey and Jared Stabach at the Smithsonian's National Zoo \& Conservation Biology Institute. We thank the annotation volunteers, Spencer Harman, Lani O'Foran, Autumn Gray, Sydney Houck, Jessica Winey, Emily Csizmadia, and Isabella Barrera, for improving our dataset quality.  We extend our gratitude to American Prairie, especially Danny Kinka and Dan Stevenson, for their support and collaboration on this project.

{
    \small
    \bibliographystyle{ieeenat_fullname}
    \bibliography{main}
}


\end{document}